\documentclass[10pt, a4paper]{article}

\usepackage{lrec-coling2024} 

\usepackage{latexsym}
\usepackage{comment}
\usepackage{svg}

\usepackage{graphicx}

\usepackage{color}

\definecolor{ForestGreen}{RGB}{34,139,34}
\definecolor{electricindigo}{rgb}{0.44, 0.0, 1.0}
\definecolor{cobalt}{rgb}{0.8, 0.28, 0.8}

\title{Linguistic Knowledge Can Enhance Encoder-Decoder Models \\(\textit{If You Let It})}

\name{Alessio Miaschi, Felice Dell'Orletta, Giulia Venturi} 

\address{ItaliaNLP Lab, Institute for Computational Linguistics "A. Zampolli" (CNR-ILC), Pisa\\
        \{name.surname\}@ilc.cnr.it}

\abstract{
In this paper, we explore the impact of augmenting pre-trained Encoder-Decoder models, specifically T5, with linguistic knowledge for the prediction of a target task. In particular, we investigate whether fine-tuning a T5 model on an intermediate task that predicts structural linguistic properties of sentences modifies its performance in the target task of predicting sentence-level complexity. Our study encompasses diverse experiments conducted on Italian and English datasets, employing both monolingual and multilingual T5 models at various sizes. Results obtained for both languages and in cross-lingual configurations show that linguistically motivated intermediate fine-tuning has generally a positive impact on target task performance, especially when applied to smaller models and in scenarios with limited data availability. 
 \\ \newline  }

\begin{document}

\maketitleabstract

\section{Introduction}
\label{introduction}

Understanding ``how linguistic concepts that were common as features in NLP systems are captured in neural networks'' \citep{belinkov-glass-2019-analysis} has been the focus of many studies in the recent NLP research.  It has been extensively shown that pre-trained Neural Language Models (NLMs) are able to capture syntax- and semantic-sensitive phenomena \citep{hewitt-manning-2019-structural,pimentel-etal-2020-information,li-etal-2022-probing-via} and that there is a correlation between the degree of linguistic knowledge and its ability to solve correctly a downstream task 
\citep{miaschi-etal-2020-linguistic,sarti-etal-2021-looks}, although it is still highly debated \citep{ravichander-etal-2021-probing}. However, it has also been demonstrated that introducing additional linguistic information \citep{wang2019structbert,zhou-etal-2020-limit,glavas-vulic-2021-supervised} during the pre-training phase can enhance models' performances. In addition, several works showed that transfer learning methods, such as fine-tuning on intermediate supporting tasks, are highly beneficial to improve pre-trained models' performance in the resolution of multiple final target tasks \citep{phang:2018,wang-etal-2019-tell}. 

\begin{figure}[t!]
\centering
\includegraphics[width=0.47\textwidth]{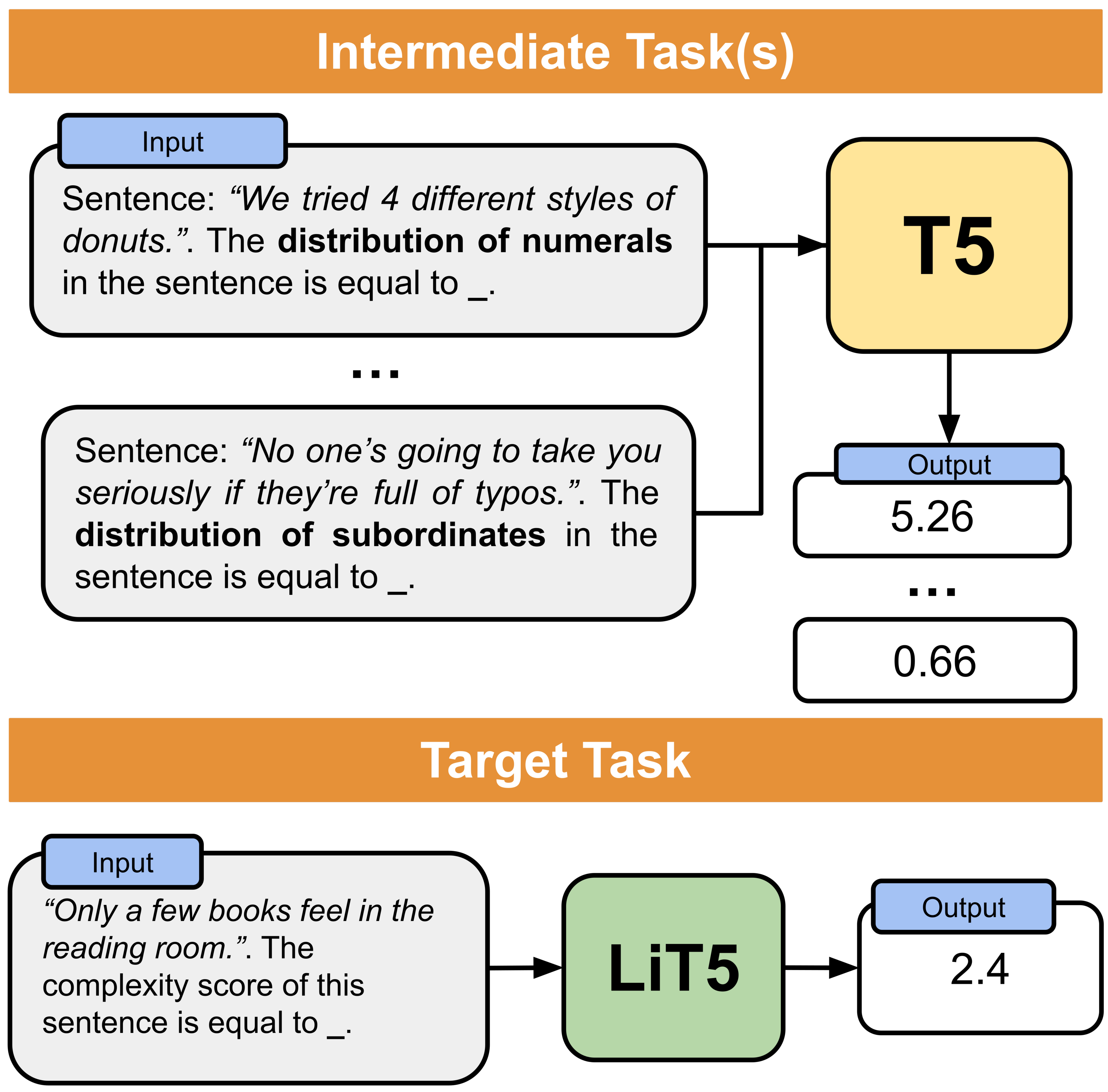}
\caption{Illustrated example of the proposed methodology. T5 is previously fine-tuned on a subset of linguistic intermediate tasks in a multitask fashion. Then, the newly obtained model, LiT5, is tested on the target task.}
\label{fig:approach}
\end{figure}

Starting from these premises, in this work we conducted multiple experiments devoted to evaluating the potential benefits of enriching a pre-trained NLM with multiple linguistic information that may enhance its performance in the resolution of a target task. Specifically, we defined a methodology based on an intermediate fine-tuning phase where the model is instructed to solve a set of raw, morpho-syntactic and syntactic tasks both in a multi- and single-task scenario. To this end, we tested our method on an Encoder-Decoder model, i.e.\ T5 \cite{raffel2020exploring}, that allows focusing the instructing process on specific linguistic tasks.
Our target task is one that strongly relies on the knowledge of the linguistic properties that characterize a sentence, namely, predicting the linguistic complexity of a sentence (see Figure \ref{fig:approach} for an overview of the methodology).

The methodology was tested on two languages, Italian and English, employing both mono- and multi-lingual T5 models. Furthermore, we devised a cross-lingual evaluation scenario to assess the effectiveness of a model that underwent linguistic fine-tuning on data from a language different from that of the target task.
The experiments on the target task were conducted by varying the training data size to explore how linguistically informing the models impacts their performance across scenarios with limited data.
Our purpose is not to propose a new framework for achieving state-of-the-art performance, but rather to inspect how this process of enhancing a model with linguistic knowledge scales across different languages and amounts of data.

Our main contributions are:
\begin{itemize}
    \item We propose an intermediate fine-tuning approach to study the impact of enhancing pre-trained Encoder-Decoder models with knowledge of multi-level linguistic phenomena in solving a target task that has not been tested so far.
    \item We compare the effectiveness of linguistically informing both mono- and multi-lingual models, highlighting language-specific peculiarities.
    \item We test our approach on T5 models of increasing size and compare their performance varying the dimensions of the target training, thus uncovering the potential for resource-efficient, linguistically-informed small models, particularly in data-limited scenarios.
    \item We demonstrate the applicability of our method of linguistic fine-tuning across languages, offering valuable adaptability insights for cross-lingual scenarios\footnote{Datasets and models are available at the following repository: \url{https://github.com/alemiaschi/linguistically_informed_t5}.}.
\end{itemize}

\section{Related Work}
\label{rel_work}

A large body of recent work focused on investigating whether performing multiple steps of fine-tuning on one or more intermediate tasks can enhance the performance of a pre-trained NLM before fine-tuning it on a target task. The core concept behind this approach is to specialize the pre-trained model by exposing it to tasks other than language modeling, which can enhance its capabilities through a transfer-learning approach. While conducting a comprehensive survey of the literature on this topic is beyond the scope of our work, we aim to concentrate on three key aspects that are frequently discussed about the effectiveness of the intermediate fine-tuning process. These aspects are interconnected, although we present them sequentially.
The first one concerns the size of the intermediate fine-tuning and target datasets, and is generally related to a second aspect concerning the fine-tuning approach adopted \citep{vu-etal-2020-exploring,chang-lu-2021-rethinking-intermediate}. Among the others, e.g.\ \citet{weller-etal-2022-use} showed that a multi-task learning approach (i.e.\ fine-tuning on a supporting task and the target task simultaneously) to the intermediate fine-tuning on the GLUE benchmark \citeplanguageresource{wang-etal-2018-glue} is more effective than conducting the two phases consecutively (a methodology often called STILTs) when the target dataset is smaller than the supporting dataset and vice-versa.

A third aspect highly discussed relates to the selection of intermediate tasks \citep{zhang-zhang-2021-qa,padmakumar-etal-2022-exploring,goot-2023-machamp}. Generally, the effectiveness of the intermediate fine-tuning process is highly dependent on the choice of these tasks. For instance, \citet{pruksachatkun-etal-2020-intermediate} performed a study on the RoBERTa model \cite{liu2019roberta} by fine-tuning it on 110 different combinations of intermediate tasks and evaluating the trained models via \textit{probing tasks}. The experiments showed that intermediate tasks requiring high-level inference and reasoning abilities tend to work best.
Differently from previous studies, our approach to intermediate fine-tuning is more `phenomenon-oriented' than `task-oriented'. Specifically, we have devised a set of prediction tasks that reflect the progressive acquisition of multiple linguistic phenomena, which could prove beneficial in resolving the target task.

On a different side, different works focused on enhancing NLMs with linguistic knowledge during their pre-training phase \cite{wang2019structbert,glavas-vulic-2021-supervised}. Among these, \citet{zhou-etal-2020-limit} enhanced a BERT model by training it in a multi-task learning technique with five syntax and semantics tasks, showing that the model can outperform a base one on several benchmarks. 

In our work, we aim to bridge these perspectives by exploring an approach that combines the enhancement of a model with linguistic knowledge through intermediate-task fine-tuning, thereby exploiting the power of both transfer learning methods and linguistic knowledge integration.

\section{Experimental Setting}
\label{approach}

We devised a two-step STILTs approach. Firstly, we fine-tuned the T5 models on several intermediate support tasks. They consist of a set of linguistic phenomena modeling aspects of a sentence related in the literature to complexity. 
They serve as input features for the models which are trained for a total of 25 epochs to predict their values in a multi-task setting. We saved model checkpoints at regular intervals, specifically every 5 epochs, thus resulting in 5 distinct snapshots of the models, each representing a different phase in their development\footnote{The LiT5 models are available at the following link: \url{https://huggingface.co/collections/alemiaschi/lit5-65d5d480dd8841806fc0a2a0}.}.
Secondly, we fine-tuned the \textit{``Linguistically informed''} T5 models, LiT5s, on the prediction of sentence complexity (one for each checkpoint), which we chose as the target task. 

The two steps were tested using Italian and English T5 
and the multilingual version. Additionally, we introduced a cross-lingual setting designed to evaluate the efficacy of multilingual models linguistically fine-tuned in a language other than that of the task. For both the intermediate fine-tuning and the target task phase, we verbalized each input sequence by adding a suffix that clarifies the task to be solved. For example, to instruct the models with linguistic knowledge we postpend each sentence with a set of verbalizations,  as in the following example (the suffix is in bold): `\textit{In 1982, he started a factory in Greece. \textbf{The distribution of numerals in the sentence is equal to \_}}''. 
While, for the target task, we postpend the string \textit{``The complexity score of the sentence is equal to \_''}\footnote{The complete list of the verbalizations can be found in Appendix \ref{sec:appendix_a}.}. 

We also proposed a further evaluation scenario to investigate which linguistic feature 
is most informative for improving the models in solving the target task. To this end, we performed the same intermediate fine-tuning process, but 
in a single-task scenario, training the small monolingual models with each feature 
at a time. 

\subsection{Data}
\label{data}

\paragraph{Target Task} We considered the task of predicting the level of linguistic complexity of a sentence and we relied on the corpus introduced in \citetlanguageresource{brunato-etal-2018-sentence}. It contains 
Italian sentences taken from the newspaper section of the Italian Universal Dependency Treebank (IUDT) \citeplanguageresource{simi-etal-2014-less} and 
English sentences extracted from the Wall Street Journal section of the Penn Treebank \citeplanguageresource{mcdonald-etal-2013-universal} manually rated by 20 crowdsourced workers for the corresponding level of perceived complexity on a 1-to-7-point scale. For our experiments, we decided to consider the average judgment of complexity given by all annotators to each sentence. For instance, the average complexity score associated with the following sentence \textit{"Only a few books feel in the reading room."} is \textit{2.4}.

All the sentences contained in the two treebanks were grouped into 6 bins based on a different sentence length, i.e.\ 10, 15, 20, 25, 30, 35 tokens. Our intention in controlling sentence length was to create sets of sentences that feature comparable values for linguistic characteristics known to be associated with sentence length, such as parse tree depth or dependency links.
The original corpus was divided in two: 50\% was used as training and 50\% as testing. We made sure to maintain a balanced distribution of sentences of 6 different lengths in order to expose the model to a diverse range of sentence lengths during training and testing. For the two languages, we further divided the training set into 5 bins containing an increasing number of sentences and balanced for the six sentence lengths. Specifically, we created 5 training sets each containing 120, 240, 360, 480, and 600 sentences for Italian and 72, 144, 216, 288, and 360 sentences for English. The test sets contain 600 and 360 sentences for Italian and English, respectively. We conducted experiments using training sets of increasing dimensions to explore the impact of enhancing the models with linguistic knowledge also in scenarios with limited data.

\paragraph{Intermediate Tasks} They consist of predicting in a multi- and single-task setting the set of linguistic features selected with the approach described in \ref{features}. 
The values of the features were extracted from version 2.5 of the Italian and English Universal Dependency (UD) treebanks \citeplanguageresource{Zeman:2019}. In total, we collected 16,000 and 4,000 sentences for the training and tests of the Italian treebank, and 17,600 and 4,400 
for the English one.

\subsection{Models}
\label{models}

We relied on different versions of the T5 model \cite{raffel2020exploring}. For the experiments devised on English, we utilized three models of increasing size, all trained on the English language: \textit{t5-small} (60M), \textit{t5-base} (220M), and \textit{t5-large} (770M). For experiments related to the Italian language, we employed IT5 \cite{sarti2022it5}, a T5 model trained on the Italian sentences extracted from a cleaned version of the mC4 corpus \cite{xue-etal-2021-mt5}. Just like with the English models, we tested IT5 in three different sizes: \textit{it5-small} (60M), \textit{it5-base} (220M), and \textit{it5-large} (738M). 

In addition to these language-specific models, we also conducted cross-lingual experiments using the multilingual model. In this regard, we tested both \textit{mt5-small} (300M) and \textit{mt5-base} (580M). We used Huggingface’s transformers library \cite{wolf-etal-2020-transformers} for accessing all the models\footnote{For details regarding models, compute parameters and training details see Appendix \ref{sec:appendix_b}.}.

\paragraph{Evaluation} We used Spearman correlation score as evaluation metric. In particular, we computed the Spearman correlation between the gold value of each feature in the Italian or English treebank and the predicted value of the models for the intermediate tasks. For what concerns the target task, we computed the correlation between average judgments of complexity and the complexity scores obtained with the fine-tuned LiT5 models.

\subsection{Linguistic Features}
\label{features}

The set of linguistic features we used as intermediate tasks consists of a subset of those extracted with the ProfilingUD \cite{brunato-etal-2020-profiling}, a tool that allows the extraction of more than 130 properties representative of the linguistic structure underlying a sentence and derived from raw, morpho-syntactic and syntactic levels of annotation based on the UD formalism \cite{10.1162/coli_a_00402}.
The key advantage of relying on the UD formalism lies in the possibility of encoding different sentence properties consistently across various languages. This choice facilitates the adaptation of our methodology to diverse languages with relative ease. Additionally, these features have been shown to play a highly predictive role when leveraged by traditional learning models on various classification problems and can be also effectively used to profile the knowledge encoded in the internal representations of a pre-trained NLM \cite{miaschi-etal-2020-linguistic}.

We select the subset of linguistic features based on the degree of correlation between sentence-level complexity judgments and the values of linguistic features 
extracted by Profiling-UD from the target task datasets. 
Since, as pointed out by \citet{brunato-etal-2018-sentence}, 
linguistic complexity is strongly correlated with sentence length, we first calculated the correlation for each of the 6 bins, each containing sentences of different lengths, then averaged the correlation scores. This step was undertaken to minimize the influence of features 
directly related to sentence length, allowing us to concentrate on linguistic characteristics that contribute more indirectly to influencing the human perception of linguistic complexity. 
In the end, we obtained a list of the most correlated features 
for each language considered regardless of sentence length. 

\paragraph{Selected features} The list of the 10 most correlated ones, along with their Spearman correlation scores, are reported in Table \ref{tab:selected_features}. A noteworthy observation is that the scores for both languages appear relatively low. This can be attributed to our deliberate exclusion of features 
associated with sentence length, which typically exhibits the highest correlations with complexity. Interestingly, for each language, we find different linguistic phenomena, covering diverse aspects of sentence complexity. They include a {\bf raw text feature} as the average length of words (\textit{char\_per\_tok}); distribution of specific {\bf parts-of-speech} (\textit{upos\_dist\_*}), i.e.\ adjectives (\textit{ADJ}), numbers (\textit{NUM}), punctuation marks (\textit{PUNCT}), auxiliary verbs (\textit{AUX}), determiners (\textit{DET}), pronouns (\textit{PRON}), symbols (\textit{SYM}), ratio of content parts-of-speech over the total number of words (\textit{lexical\_density}), of {\bf syntactic dependency relations} (\textit{dep\_dist\_*}), i.e.\ function words associated with a verb (\textit{aux}), markers introducing a subordinative clause (\textit{mark}), multiword expressions (\textit{compound}), numeric modifiers of a noun (\textit{nummod}), of auxiliary verbs by {\bf inflectional morphology traits} (\textit{aux\_*}), i.e.\ finite forms (\textit{form\_dist\_Fin}), indicative moods (\textit{mood\_dist\_Ind}), and of {\bf syntactic tree structures}, i.e.\ direct objects in post-verbal position (obj\_post) and subordinate clauses (\textit{subord\_prop\_dist})\footnote{A detailed description of the selected linguistic features is available in Appendix \ref{sec:appendix_c}.}.

\begin{table}[t!]
\centering
\scriptsize
\begin{tabular}{lrlr}
\hline
\textbf{Features}                       & \textbf{Corr} & \textbf{Features}                       & \textbf{Corr} \\
\hline
\multicolumn{2}{l}{\textbf{Italian}}                  & \multicolumn{2}{l}{\textbf{English}}                  \\
\hline
char\_per\_tok                 & 0.28        & upos\_dist\_NUM                & 0.35        \\
upos\_dist\_ADJ                & 0.21        & dep\_dist\_nummod              & 0.31        \\
upos\_dist\_NUM                & 0.19        & upos\_dist\_SYM                & 0.27        \\
lexical\_density               & 0.17        & upos\_dist\_AUX                & 0.25        \\
dep\_dist\_aux                 & 0.17        & dep\_dist\_compound            & 0.25        \\
dep\_dist\_mark                & 0.16        & upos\_dist\_PRON               & 0.24        \\
aux\_mood\_dist\_Ind           & 0.14        & upos\_dist\_DET                & 0.23        \\
obj\_post                      & 0.14        & subord\_prop\_dist & 0.17        \\
upos\_dist\_PUNCT              & 0.13        & aux\_form\_dist\_Fin           & 0.16        \\
subord\_prop\_dist & 0.12        & aux\_mood\_dist\_Ind           & 0.14       \\
\hline
\end{tabular}
\caption{Linguistic features selected as intermediate tasks along with their average correlation score (Spearman $\rho$ coefficient) with the complexity judgments. 
All scores are statistically significant ($p-value<0.05$).}
\label{tab:selected_features}
\end{table}

\section{Results}
\label{results}

In the following sections, we delve into the outcomes of our experiments. First, we present the results of the T5 models on the intermediate tasks. Subsequently, we analyze the results of the target task, comparing the performance of both the base and LiT5 models.

\subsection{Enhancing T5 with Linguistic Features}

\begin{figure*}[t!]
\centering
\includegraphics[width=0.95\textwidth]{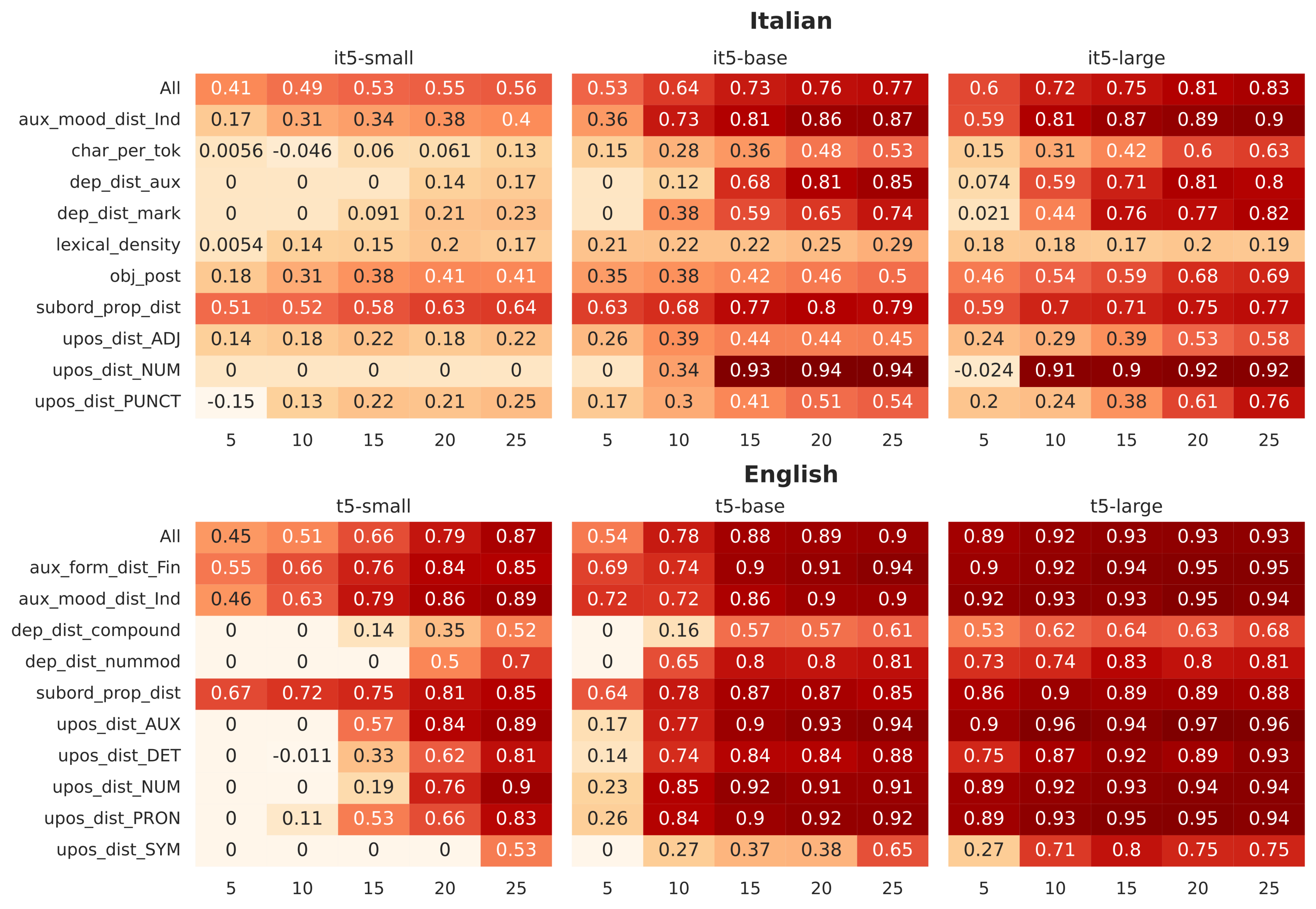}
\caption{Spearman correlation coefficients for the intermediate tasks for the Italian (\textit{top}) and English (\textit{bottom}) datasets obtained with the monolingual T5 models. Each column in the heatmaps contains the results obtained by the models fine-tuned for multiple epochs (e.g. 5 = 5 epochs of 
fine-tuning).}
\label{fig:intermediate_heatmap}
\end{figure*}

\paragraph{Monolingual Models} Figure \ref{fig:intermediate_heatmap} reports the results for the intermediate task (in terms of Spearman correlation coefficients) obtained with the monolingual \textit{t5} models in a multi-task learning scenario.  
As a first observation, we can see that for both Italian and English, {\bf all models tend to become progressively ``linguistically-informed'' over the 25 epochs of fine-tuning}. 

Interestingly, it seems that {\bf the size of the model impacts the results of the fine-tuning process}. On the one hand, large models consistently outperform smaller ones;
on the other hand, it's noteworthy that the improvement over epochs is particularly pronounced for smaller models. This observation possibly suggests that weights of smaller models even trained on the same amount of data and for the same number of epochs are more susceptible to modification. In addition, we can assume that smaller pre-trained models implicitly encode less linguistic knowledge than larger ones, thus making the addition of new explicit linguistic information more effective. 

When we focus on the ranking of each learned linguistic feature, we can generally observe that there are multiple distinctions across models of different sizes. 
Specifically, it results that at the end of the fine-tuning process the \textit{it5-small} model excels in mastering the distribution of subordinate clauses (\textit{subord\_prop\_dist}) in comparison with other features learned by the same model, while the distribution of numerals (\textit{upos\_dist\_NUM}) is the best-mastered property by \textit{it5-base} and \textit{large}. 
Conversely, the prediction of values of the lexical density of a sentence and of token length (\textit{char\_per\_tok}) remains consistently a challenge: they are among the worst-mastered sentence characteristics for all the models.

Interestingly, model size also seems to have an impact on the learning speed of specific features. 
For example, the difference in the accuracy obtained between after 5 and 10 fine-tuning epochs in predicting the lexical density is much greater for \textit{it5-small} than for larger models, for which we obtained quite similar 
scores across epochs. 
A quite peculiar case is represented by numbers (\textit{upos\_dist\_NUM}): \textit{it5-small} fails to master them even at the end of the entire fine-tuning process. 

Similar observations hold for \textit{t5}: the model size has an impact on the performance, even if {\bf English models consistently outperform the Italian ones}. However, regardless of their dimensions, all three English models exhibit the lowest proficiency considering the same three features: 
the distribution of numerical modifiers (\textit{dep\_dist\_nummod}), compounds (\textit{dep\_dist\_compound}), and symbols (\textit{upos\_dist\_SYM}). When we focus on the learning speed across epochs, we can observe that information about the distribution of syntactic dependencies (\textit{dep\_dist\_*}) is acquired by \textit{t5-small} when it is fine-tuned for 10 epochs. Conversely, \textit{t5-base} and \textit{large} master these features earlier and consistently achieve superior results, in line with the general trend already noted.  
The prediction of symbols (\textit{upos\_dist\_SYM}) represents a quite peculiar case: they start to be predicted by \textit{t5-small} only after 20 epochs, and it is among the worst-predicted features by the bigger models.
\begin{figure}[t!]
\centering
\includegraphics[width=0.49\textwidth]{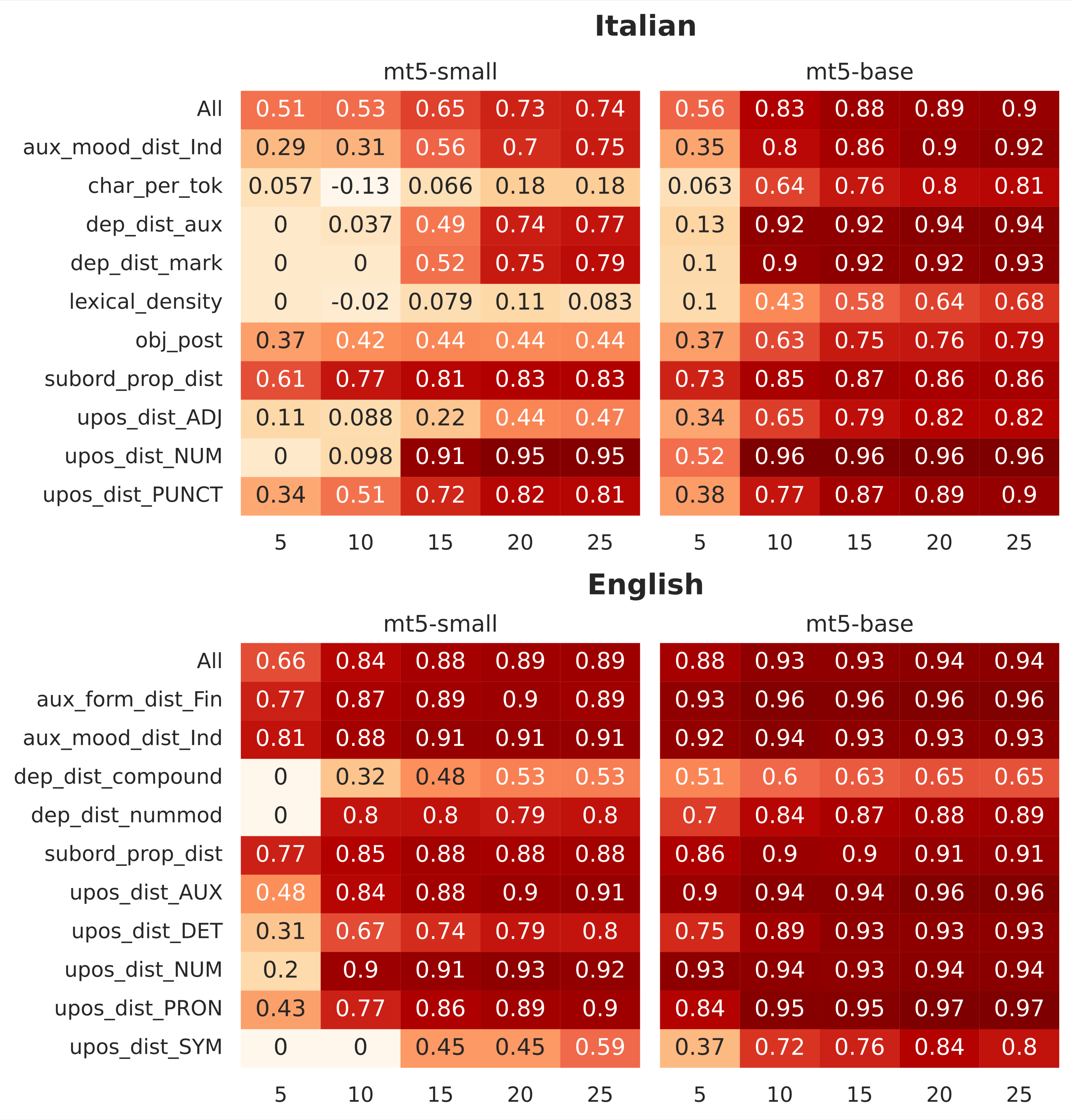}
\caption{Spearman correlation coefficients for the intermediate tasks for the Italian (\textit{top}) and English (\textit{bottom}) datasets obtained with the multilingual T5 models.}
\label{fig:intermediate_heatmap_multi}
\end{figure}

\paragraph{Multilingual Models} As we can observe from Figure \ref{fig:intermediate_heatmap_multi}, {\bf multilingual models consistently outperform monolingual ones} for both languages.
In addition, some differences in terms of learning speed can be highlighted. In particular, for all features, the improvement of \textit{mt5-base} fine-tuned on the Italian treebank for 10 epochs is much higher than for \textit{it5-base} trained for the same numbers of epochs: this suggests that it becomes more rapidly 
informed on the same linguistic characteristics 
than the monolingual model. In general, the small model undergoes the main differences compared to its monolingual counterpart.

Similar differences in the learning speed can be highlighted for the English language, especially in the case of the small model. For example, symbols (\textit{upos\_dist\_SYM}) are learned only after 20 epochs by \textit{t5-small} but after 10 by \textit{mt5-small}, or numerical modifiers (\textit{upos\_dist\_NUM}) after 15 epochs by the monolingual model but only after 5 by the multilingual counterpart.

\subsection{Predicting Complexity with Linguistically-Informed Models}
\label{predicting_complexity}

\begin{figure*}[t!]
\centering
\includegraphics[width=1\textwidth]{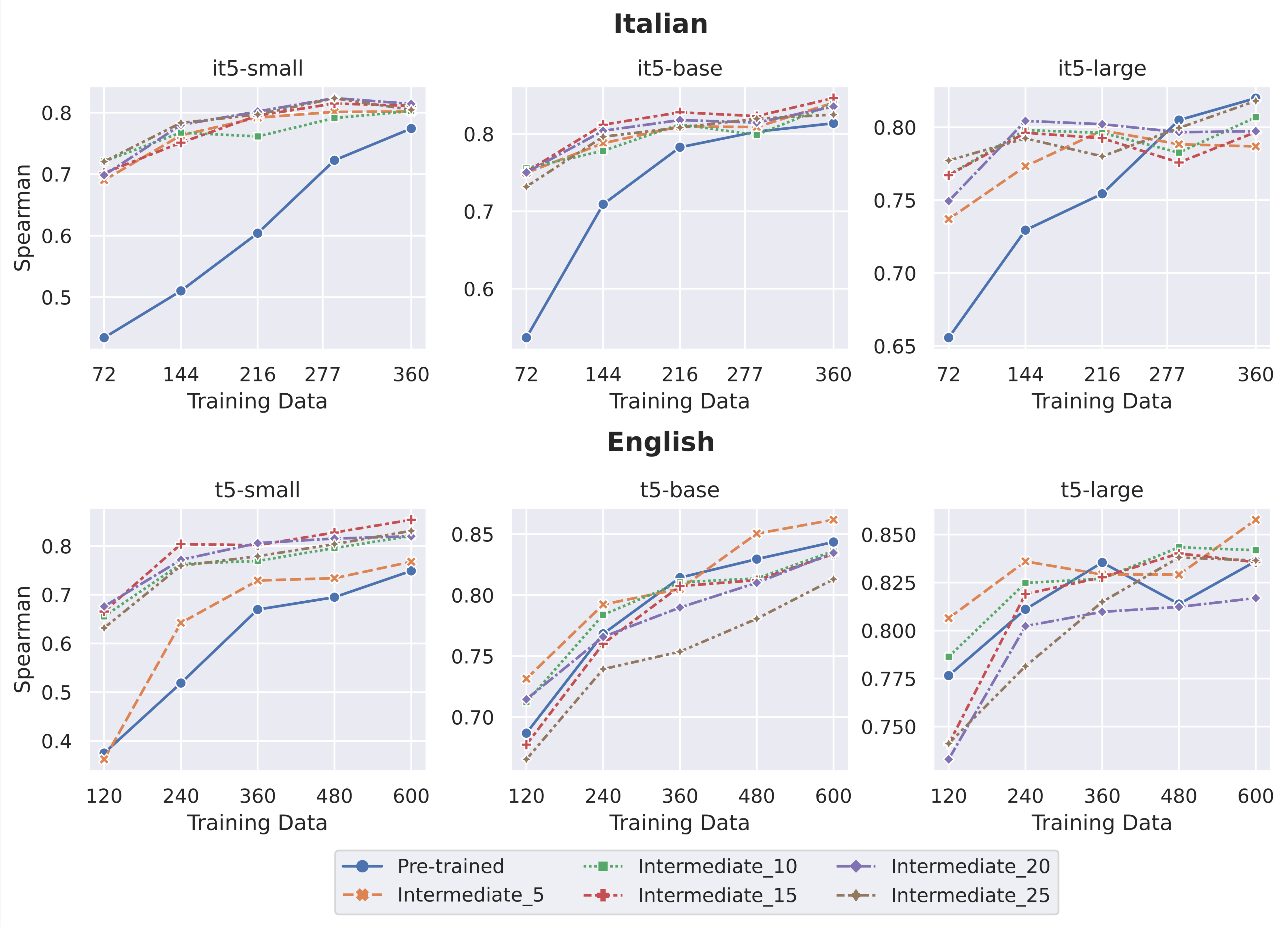}
\caption{Spearman correlation coefficients for the target tasks for the Italian (\textit{top}) and English (\textit{bottom}) datasets obtained with the monolingual models using pre-trained and LiT5 models.}
\label{fig:target_scores}
\end{figure*}

\paragraph{Monolingual Models} In Figure \ref{fig:target_scores} we report the performance trend of the \textit{Pre-Trained} and the LiT5 models  
when fine-tuned on the different portions of the target dataset and for increase numbers of epochs (\textit{Intermediate\_*}).

The first notable observation is the varying trends observed across model dimensions and languages. {\bf For the two languages, smaller models consistently outperform their pre-trained counterparts}, exhibiting a more pronounced impact from intermediate fine-tuning. This effect becomes even {\bf more evident when working with limited training data}, highlighting the advantages of linguistic knowledge 
enhancement through a step of intermediate fine-tuning. Particularly evident is, in fact, the performance gap when fine-tuning the models with a small portion of the training dataset. 
For instance, when the models are fine-tuned using only 1/5 of the entire training dataset, the disparities in performance of \textit{it5-small} and \textit{t5-small} are notably pronounced (0.29 and 0.30 between the \textit{Pre-trained} and LiT5 correlation scores respectively). 
In contrast, when {\bf considering bigger models (base and large), we notice different behavior across languages}. For the Italian language, LiT5s outperform the pre-trained ones when fine-tuned on the first three portions of the dataset. However, their performance tends to converge with that of the pre-trained models as the fine-tuning process utilizes the majority of the available training data. In contrast, the trends for the English portion of the dataset are more nuanced, with variations between base and large models and distinctions emerging across different intermediate training epochs. These findings underscore the intricate interplay between model size, language, and the efficacy of intermediate fine-tuning in optimizing performance for specific tasks and datasets.

Focusing instead on the distinctions among the various LiT5 models, we uncover varying trends, particularly when considering different languages. For Italian, there appears to be no discernible difference among models or a consistent trend across increasing training sizes and model dimensions. In fact, scores among the LiT5 models remain generally quite similar.  
More pronounced variations are exhibited by the \textit{it5-large} model: when trained for only 5 epochs (\textit{Intermediate\_5}) it yields generally lower scores compared to its counterpart trained for 25 epochs (\textit{Intermediate\_25}). 

In contrast, surprisingly, {\bf English smaller models benefit most from extended intermediate fine-tuning processes (> 5 epochs), whereas the base and large models achieve their best performance when linguistically enhanced for only 5 epochs.} This trend becomes especially evident when evaluating performance on the entire target dataset, where the LiT5 models outperform the pre-trained ones. 
A possible explanation of this trend is that a smaller model may require a longer intermediate fine-tuning process to better adapt to the task, by gradually acquiring the necessary linguistic knowledge. 
On the other hand, larger models have already learned a wider range of linguistic features during their pre-training
, reducing the need for extensive fine-tuning. This transfer learning advantage allows them to quickly adapt to specific tasks and perform well with shorter fine-tuning processes.

\begin{figure}[t!]
\centering
\includegraphics[width=0.48\textwidth]{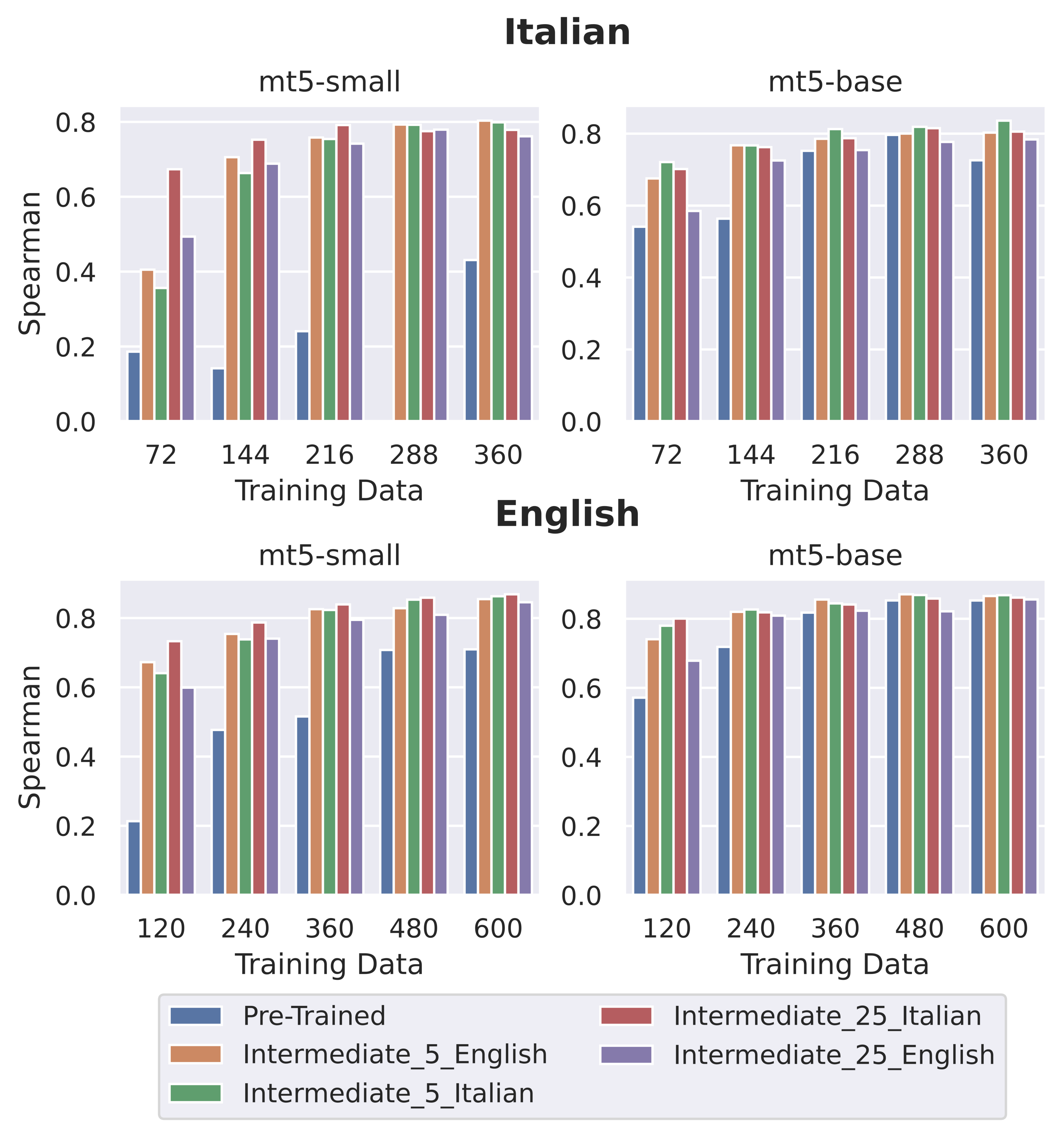}
\caption{Spearman correlation coefficients for the target tasks for the Italian (\textit{top}) and English (\textit{bottom}) datasets obtained with the \textit{mt5-*}. 
Intermediate scores are reported for the models fine-tuned for Italian and English for the minimum (5) and maximum (25) number of epochs.}
\label{fig:target_scores_multi}
\end{figure}

\paragraph{Multilingual Models} Figure \ref{fig:target_scores_multi} shows the performance trend of the \textit{Pre-Trained} and the LiT5 multilingual models\footnote{We reported only the results obtained by the LiT5s fine-tuned for the minimum (5) and maximum (25) number of epochs. Complete results are reported in Appendix \ref{sec:appendix_d}.}. 
As it can be noted, there is a similar trend to what we observed with monolingual models. {\bf In most configurations, multilingual LiT5 models intermediately fine-tuned either on Italian (\textit{*\_Italian}) and English (\textit{*\_English}) improve the performance on the target task, especially for smaller models and with limited data available.}
If we focus instead more on the various LiT5 models, we notice that, in general, when training data is limited (1/5 or 2/5 of the entire dataset) the most significant impact for small models is seen in those that have completed the intermediate fine-tuning cycle (25 epochs), while training with larger portions of the dataset tends to reduce the difference between the models. 
In contrast, for \textit{mt5-base} this tendency is more attenuated: the improvement is evident even when models are trained with only 1/5 of the dataset. 

Additionally, we can see that, regardless of whether the LiT5 multilingual models were obtained through fine-tuning on a language different from the target task, their impact is still evident, suggesting the {\bf effectiveness of the intermediate tuning even in cross-lingual configurations}. However, it is interesting to notice a substantial difference between the target tasks in the two languages. In fact, if we look at the results obtained on the Italian target task, we can see that having performed intermediate tasks in the same language as the target tends to be more effective than doing it in English. On the other hand, a different trend emerges with the English dataset: performing intermediate tasks on Italian data 
tends to increase the performance more than using a linguistically-informed model trained on English data. 
In fact, the performance of the LiT5 small model tested on Italian data and intermediately trained on English data increases by 0.31 points in terms of Spearman scores, while it increases by 0.52 when tested on English data and further trained on Italian data. Similar differences are observed for the base models, with a difference of 0.13 and 0.25 in the two scenarios\footnote{
Further details about performance differences are available in Appendix \ref{sec:appendix_d}.}.
This result seems to suggest that the enrichment of a multilingual model with language-specific linguistic knowledge 
is more effective for a language that is poorly represented in the original training of a multilingual model, such as Italian\footnote{The Italian language covers around 2.43\% of the mC4 corpus \cite{xue-etal-2021-mt5}.}. In contrast, a more represented language such as English, which might be widely informed of language-specific information, benefits more from an intermediate training 
on a different language.

\subsection{Which linguistic features matter most?}
\label{impact_feature}

\begin{table}
\scriptsize
\centering
\begin{tabular}{lrlr}
\hline
\textbf{Features}                        & \textbf{Corr} & \textbf{Features}                        & \textbf{Corr} \\
\hline
\multicolumn{2}{l}{\textbf{Italian}} & \multicolumn{2}{l}{\textbf{English}} \\
\hline
upos\_dist\_NUM                & 0.65     & aux\_mood\_dist\_Ind           & 0.92     \\
subord\_prop\_dist & 0.60     & aux\_form\_dist\_Fin           & 0.86     \\
aux\_mood\_dist\_Ind           & 0.52     & subord\_prop\_dist & 0.81     \\
obj\_post                      & 0.48     & upos\_dist\_PRON               & 0.80     \\
upos\_dist\_ADJ                & 0.22     & upos\_dist\_NUM                & 0.65     \\
char\_per\_tok                 & 0.18     & dep\_dist\_nummod              & 0.34     \\
lexical\_density               & 0.13     & upos\_dist\_DET                & 0.24     \\
upos\_dist\_PUNCT              & 0.09     & upos\_dist\_AUX                & -0.03    \\
dep\_dist\_mark                & -0.03    & dep\_dist\_compound            & -0.03    \\
dep\_dist\_aux                 & -0.03    & upos\_dist\_SYM                & $\#\#$   \\
\hline
\end{tabular}
\caption{Spearman correlation coefficients for the intermediate tasks for the Italian and English datasets obtained with the small models fine-tuned with one linguistic feature at a time.}
\label{table:intermediate_single}
\end{table}

To have a deeper understanding of the contribution of each linguistic feature 
to the target task, we propose a final evaluation by testing the Italian and English monolingual small models (\textit{it5-small} and \textit{t5-small})  when previously fine-tuned on each of the 10 selected linguistic features 
at a time (i.e. one LiT5 model per property). Note that for this scenario, we have selected only one LiT5 model per feature\footnote{We kept the models that maximize the performance in the prediction of the linguistic features.} instead of testing different checkpoints, as in Sec. \ref{predicting_complexity}. First of all, we report in Table \ref{table:intermediate_single} the results for the single-feature 
intermediate tasks. Overall, we can notice that the linguistic features 
are mastered by the two models with lower precision with respect to the multi-task learning setting of fine-tuning. This observation underscores the potential advantage of jointly predicting these sentence characteristics, potentially amplifying their interoperability and mutual reinforcement.

\begin{figure}[t!]
\centering
\includegraphics[width=0.49\textwidth]{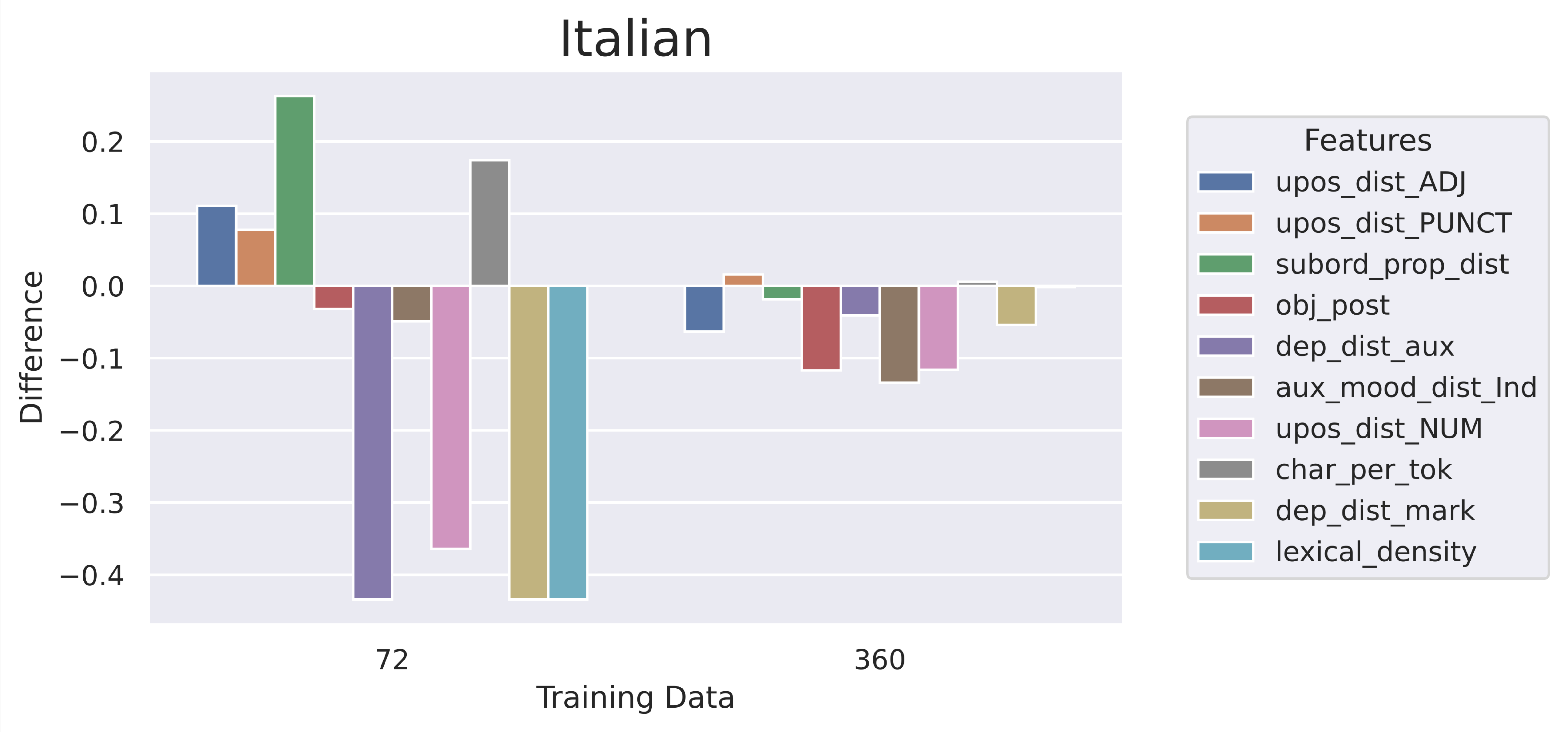}
\includegraphics[width=0.49\textwidth]{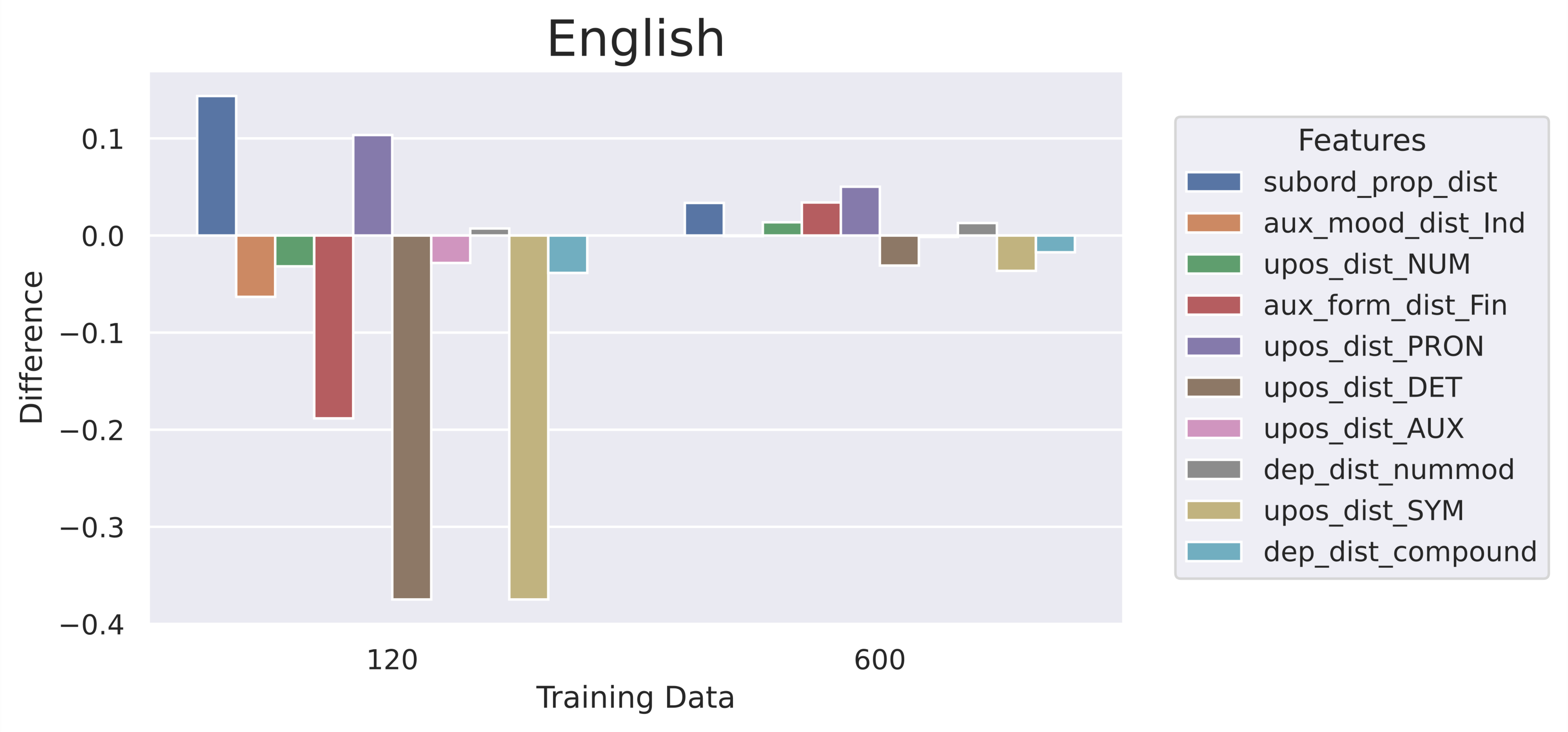}
\caption{Performance differences between the \textit{Pre-Trained} and the LiT5 models (trained on one linguistic feature at a time) in both scenarios of limited (1/5) and full training data (5/5) availability.}
\label{fig:target_single_feat}
\end{figure}

In Figure \ref{fig:target_single_feat} we report the performance differences between the \textit{Pre-Trained} and the LiT5 models in both scenarios of limited (1/5) and full training data (5/5) availability. 
The majority of negative differences indicate that, differently from the multi-task configuration, {\bf only a few features effectively help the models in the resolution of the target task}. However, the impact varies depending on the language 
and the amount of training data. In the scenario {\bf with limited resources, a higher number of features 
seem to be more informative for the Italian than for the English model and appear to contribute more significantly to enhancing the model performance} in the resolution of the target task. The two most effective ones for \textit{it5} are a syntactic feature, 
i.e.\ the use of subordinative clauses (\textit{subord\_prop\_dist}), and a raw text one, i.e.\ the length of tokens (\textit{char\_per\_tok}).
Interestingly, being informed about the presence of subordinative clauses enhances also the ability of the English model to predict the complexity of a sentence. 
Conversely, when more training data is available, the addition of linguistic information does not improve the Italian LiT5 models significantly, while it shows a slightly more beneficial effect for the English ones.

\section{Conclusion}
\label{conclusion}

In this work, we study the effectiveness of enhancing pre-trained T5 models Encoder-Decoder models (i.e.\ T5), with linguistic knowledge for the prediction of sentence complexity. We conducted a series of experiments on Italian and English datasets using both monolingual and multilingual T5 models of various sizes. Our findings reveal several key insights.

First of all, we showed that informing models linguistically over several epochs allows them to progressively improve their degree of language proficiency, albeit with some differences between languages and model size. Interestingly, model size has an impact on the speed of learning specific linguistic features.  

For what concerns the resolution of the target task, we found that our method of linguistic enhancement is 
particularly effective especially when applied to smaller models and in scenarios with limited 
availability of target training data. Interestingly, we observe that a linguistically-informed small model, refined through intermediate fine-tuning, can frequently surpass the performance of a larger model that has not undergone this intermediate refinement process. This highlights the potential for more resource-efficient models when linguistic knowledge is incorporated. This holds both for mono and multilingual models and when the linguistic fine-tuning process is conducted in a language other than that of the target task. The findings indicate that refining smaller models through efficient intermediate tuning phases could present a promising approach to building more sustainable models. Specifically, it may confer benefits compared to larger models trained for extended periods on more extensive datasets with a higher parameter count.

Finally, preliminary experiments conducted by investigating the impact of training linguistically-informed models with a single linguistic feature, highlighted that only a few of them seem to enhance the model performances. 
This result, on the one hand, underscores the evident benefits of intermediate fine-tuning within a multi-task framework. On the other hand, it poses questions for future research, emphasizing the need for a more comprehensive exploration of the 
relative impact of specific linguistic properties. 

Looking ahead, there are several avenues for future research, as detailed in the Limitations section. Additionally, it could be interesting to see if such an approach proves useful with generative Large Language Models (LLMs) in zero- and few-shot configurations. In fact, introducing linguistic knowledge via an instruction-tuning phase might enhance model performance, especially in tasks where linguistic competence plays a crucial role.

\section*{Limitations}

In this section, we discuss the limitations of our work. 1) \textbf{Languages and Tasks.} The objective of our work was to investigate if enhancing a pre-trained model with linguistic knowledge has a positive impact on its performance. However, it's important to note that while the primary focus of our work was not to propose a challenging model that performs better on multiple tasks and benchmarks, testing it on two languages and a single task could be considered a limitation. 2) \textbf{Different LLMs.} For our experiments, we primarily focused on models with "moderate" sizes, therefore more efficient and less computationally expensive. Furthermore, we selected T5 since we precisely knew its pre-training dataset. This choice was made in order to mitigate potential concerns about the incorporation of information during the intermediate tuning task that might have already been present in the model's training data. Nevertheless, we acknowledge the relevance of comparing various models in terms of architecture (e.g. encoder- or decoder-only models) and dimensions (e.g. > 1B parameters). 3) \textbf{Linguistic features.} Selecting the top 10 linguistic properties on the basis of the correlation with complexity scores is only one of the possible approaches to identifying the set of characteristics to be used for fine-tuning the model. Therefore, using a different selection method or a wider set of linguistic properties could provide more insights on how to complement the model's linguistic knowledge with additional information.

\section*{Acknowledgments}
This work has been supported by FAIR - Future AI Research (PE00000013) and XAI-CARE - 
PNRR-MAD-2022-12376692 projects under the NRRP MUR program funded by the NextGenerationEU.

\nocite{*}
\section{Bibliographical References}\label{sec:reference}

\bibliographystyle{lrec-coling2024-natbib}
\bibliography{biblio}

\section{Language Resource References}
\label{lr:ref}
\bibliographystylelanguageresource{lrec-coling2024-natbib}
\bibliographylanguageresource{languageresource}

\clearpage
\appendix

\section{Intermediate Tasks Suffixes}
\label{sec:appendix_a}

\begin{table*}[t!]
\footnotesize
\centering
\begin{tabular}{lp{12.5cm}}
\hline
\textbf{Properties}                     & \textbf{Suffixes}                                                                                                                                                  \\
\hline
char\_per\_tok                 & Il numero medio di caratteri per token nella frase è uguale a \textless{}extra\_id\_0\textgreater{} (transl. The average number of characters per token in the sentence is ...).                                                      \\
upos\_dist\_ADJ                & La distribuzione di aggettivi nella frase è uguale a \textless{}extra\_id\_0\textgreater{} (transl. The distribution of adjectives in the sentence is ...).                                                               \\
upos\_dist\_NUM                & La distribuzione dei numerali nella frase è uguale a \textless{}extra\_id\_0\textgreater{} (transl. The distribution of numbers in the sentence is ...).                                                               \\
lexical\_density               & Il rapporto fra parole piene e tutte le parole della frase è uguale a \textless{}extra\_id\_0\textgreater{} (transl. The ratio of content words over all words in the sentence is ...).                                              \\
dep\_dist\_aux                 & La distribuzione dei verbi ausiliari nella frase è uguale a \textless{}extra\_id\_0\textgreater{} (transl. The distribution of auxiliary verbs in the sentence is ...).                                                        \\
dep\_dist\_mark                & La distribuzione di marcatori che introducono una clausola subordinata ad un'altra clausola nella frase è uguale a \textless{}extra\_id\_0\textgreater{}. (transl. The distribution of markers introducing a subordinative clause to another clause in the sentence is ...)\\
aux\_mood\_dist\_Ind           & La distribuzione di verbi ausiliari all'indicativo nella frase è uguale a \textless{}extra\_id\_0\textgreater{} (transl. The distribution of indicative mood auxiliary verbs in the sentence is ...).                                          \\
obj\_post                      & La distribuzione dei verbi in posizione postverbale nella frase è uguale a \textless{}extra\_id\_0\textgreater{} (transl. The distribution of direct objects in a post-verbal position in the sentence is ...).                                         \\
upos\_dist\_PUNCT              & La distribuzione della punteggiatura nella frase è uguale a \textless{}extra\_id\_0\textgreater{} (transl. The distribution of punctuation in the sentence is ...).                                                        \\
subord\_prop\_dist & La distribuzione delle subordinate nella frase è uguale a \textless{}extra\_id\_0\textgreater{}. (transl. The distribution of subordinates in the sentence is ...) \\
\hline
\end{tabular}
\caption{The suffixes used in the multi-task intermediate fine-tuning experiments for the Italian sentences.}
\label{table:suffixes_ita}
\end{table*}

\begin{table*}[t!]
\footnotesize
\centering
\begin{tabular}{lp{12.5cm}}
\hline
\textbf{Properties}                     & \textbf{Suffixes}                                                                                   \\
\hline
aux\_form\_dist\_Fin           & The distribution of finite form auxiliary verbs in the sentence is \textless{}extra\_id\_0\textgreater{}.  \\
aux\_mood\_dist\_Ind           & The distribution of indicative mood auxiliary verbs in the sentence is \textless{}extra\_id\_0\textgreater{}. \\
dep\_dist\_compound            & The distribution of compounds in the sentence is \textless{}extra\_id\_0\textgreater{}.                    \\
dep\_dist\_nummod              & The distribution of numerical modifiers in the sentence is \textless{}extra\_id\_0\textgreater{}.          \\
upos\_dist\_AUX                & The distribution of auxiliary verbs in the sentence is \textless{}extra\_id\_0\textgreater{}.              \\
upos\_dist\_DET                & The distribution of determiners in the sentence is \textless{}extra\_id\_0\textgreater{}.                  \\
upos\_dist\_NUM                & The distribution of numerals in the sentence is \textless{}extra\_id\_0\textgreater{}.                     \\
upos\_dist\_PRON               & The distribution of pronouns in the sentence is \textless{}extra\_id\_0\textgreater{}.                     \\
upos\_dist\_SYM                & The distribution of symbols in the sentence is \textless{}extra\_id\_0\textgreater{}.                      \\
subord\_prop\_dist & The distribution of subordinates in the sentence is \textless{}extra\_id\_0\textgreater{}.     \\
\hline
\end{tabular}
\caption{The suffixes used in the multi-task intermediate fine-tuning experiments for the English sentences.}
\label{table:suffixes_eng}
\end{table*}

As mentioned in Sec. \ref{approach}, we define a set of specific suffixes to be postponed to each input sentence in the Italian/English datasets for the intermediate tasks. Tables \ref{table:suffixes_ita} and \ref{table:suffixes_eng} report the suffixes used in the experiments.

\section{Model, Parameters and Training Details}
\label{sec:appendix_b}

We fine-tuned the T5 models with the following hyperparameters:
\begin{itemize}
    \item Learning rate: 4e-5;
    \item Per device batch size: 4;
    \item Epochs: 20 epochs maximum for the complexity prediction task and 25 for the intermediate tasks. 
\end{itemize}

We trained all the models on two NVIDIA GeForce RTX 4090 GPUs. 

\section{Linguistic Properties}
\label{sec:appendix_c}

The first step of our approach consists of enhancing the T5 models with multiple linguistic properties in an intermediate fine-tuning process. For the experiments conducted on the Italian datasets we considered the following properties which can be grouped into five main groups:

\vspace{0.3cm}
\noindent Raw text feature:
\begin{itemize}
    \item \textit{char\_per\_tok}: the {\bf length of tokens} computed as the average number of characters per word in a sentence (excluding punctuation).
\end{itemize}

\noindent Distribution of morpho-syntactic categories:
\begin{itemize}
    \item \textit{ADJ}: percentage distribution of {\bf adjectives} over the total amount of tokens in a sentence according to the UD parts-of-speech tagset\footnote{https://universaldependencies.org/u/pos/index.html};
    \item \textit{NUM}: percentage distribution of {\bf numerals} over the total amount of tokens in a sentence according to the UD parts-of-speech tagset;
    \item \textit{PUNCT}: percentage distribution of {\bf punctuation marks} over the total amount of tokens in a sentence according to the UD parts-of-speech tagset;
    \item \textit{lexical\_density}: ratio of content parts-of-speech (verbs, nouns, adjectives and adverbs) over the total number of words in a sentence.
\end{itemize}

\noindent Distribution of verbs by inflectional morphology traits:
\begin{itemize}
    \item \textit{aux\_form\_dist\_Fin}: percentage distribution of  auxiliary verbs (also including modal verbs) in {\bf finite form} over the total number of auxiliary forms in a sentence according to the UD tagset\footnote{https://universaldependencies.org/u/feat/VerbForm.html};
    \item \textit{aux\_mood\_dist\_Ind}: percentage distribution of auxiliary verbs (also including modal verbs) in {\bf indicative mood} over the total number of auxiliary moods in a sentence according to the UD tagset\footnote{https://universaldependencies.org/u/feat/Mood.html}.
\end{itemize}

\noindent Distribution of syntactic dependency relations:
\begin{itemize}
    \item \textit{aux}: percentage distribution of {\bf auxiliary verbs} (also including modal verbs) over the total amount of dependency relations in a sentence according to the UD tagset\footnote{https://universaldependencies.org/u/dep/index.html};
    \item \textit{mark}: percentage distribution of {\bf markers}, i.e.\ words introducing a finite clause subordinate to another clause, over the total amount of dependency relations in a sentence according to the UD tagset.
\end{itemize}

\noindent Property referring to local syntactic tree structure:
\begin{itemize}
    \item \textit{obj\_post}: percentage distribution of direct objects that occur in a post-verbal position in a sentence.
\end{itemize}

\noindent Property referring to the use of subordination:
\begin{itemize}
    \item \textit{subord\_prop\_dist}: distribution of subordinate clauses as defined in the UD scheme \footnote{https://universaldependencies.org/u/overview/complex-syntax.html\#subordination}.
\end{itemize}

For the experiments conducted on the English datasets, we considered the following sets of properties:

\vspace{0.3cm}

\noindent Distribution of morpho-syntactic categories:
\begin{itemize}
    \item \textit{AUX}: percentage distribution of {\bf auxiliary verbs} over the total amount of tokens in a sentence according to the UD parts-of-speech tagset;
    \item \textit{DET}: percentage distribution of {\bf determiners}, i.e.\ word, also including definite and indefinite articles, that modify nouns or noun phrases, over the total amount of tokens in a sentence according to the UD parts-of-speech tagset;
    \item \textit{NUM}: percentage distribution of {\bf numerals} over the total amount of tokens in a sentence according to the UD parts-of-speech tagset;
    \item \textit{PRON}: percentage distribution of {\bf pronouns} over the total amount of tokens in a sentence according to the UD parts-of-speech tagset;
    \item \textit{SYM}: percentage distribution of {\bf symbols}, i.e.\ word-like entity that differs from ordinary words by form, function, or both, over the total amount of tokens in a sentence according to the UD parts-of-speech tagset.
\end{itemize}

\noindent Distribution of verbs by inflectional morphology traits:
\begin{itemize}
    \item \textit{aux\_form\_dist\_Fin}: percentage distribution of  auxiliary verbs (also including modal verbs) in {\bf finite form} over the total number of auxiliary forms in a sentence according to the UD tagset;
    \item \textit{aux\_mood\_dist\_Ind}: percentage distribution of auxiliary verbs (also including modal verbs) in {\bf indicative mood} over the total number of auxiliary moods in a sentence according to the UD tagset.
\end{itemize}

\noindent Distribution of syntactic dependency relations:
\begin{itemize}
   \item \textit{compound}: percentage distribution of {\bf compounds}, including all categories of multi-word expressions, over the total amount of dependency relations in a sentence according to the UD tagset;
    \item \textit{nummod}: percentage distribution of {\bf numeric modifiers}, i.e.\  number phrases that serve to modify the meaning of the noun with a quantity, over the total amount of dependency relations in a sentence according to the UD tagset.
\end{itemize}

\noindent Property referring to the use of subordination:
\begin{itemize}
    \item \textit{subord\_prop\_dist}: distribution of subordinate clauses as defined in the UD scheme.
\end{itemize}

\section{Multilingual Models Results}
\label{sec:appendix_d}

\begin{figure*}[t!]
\centering
\includegraphics[width=0.8\textwidth]{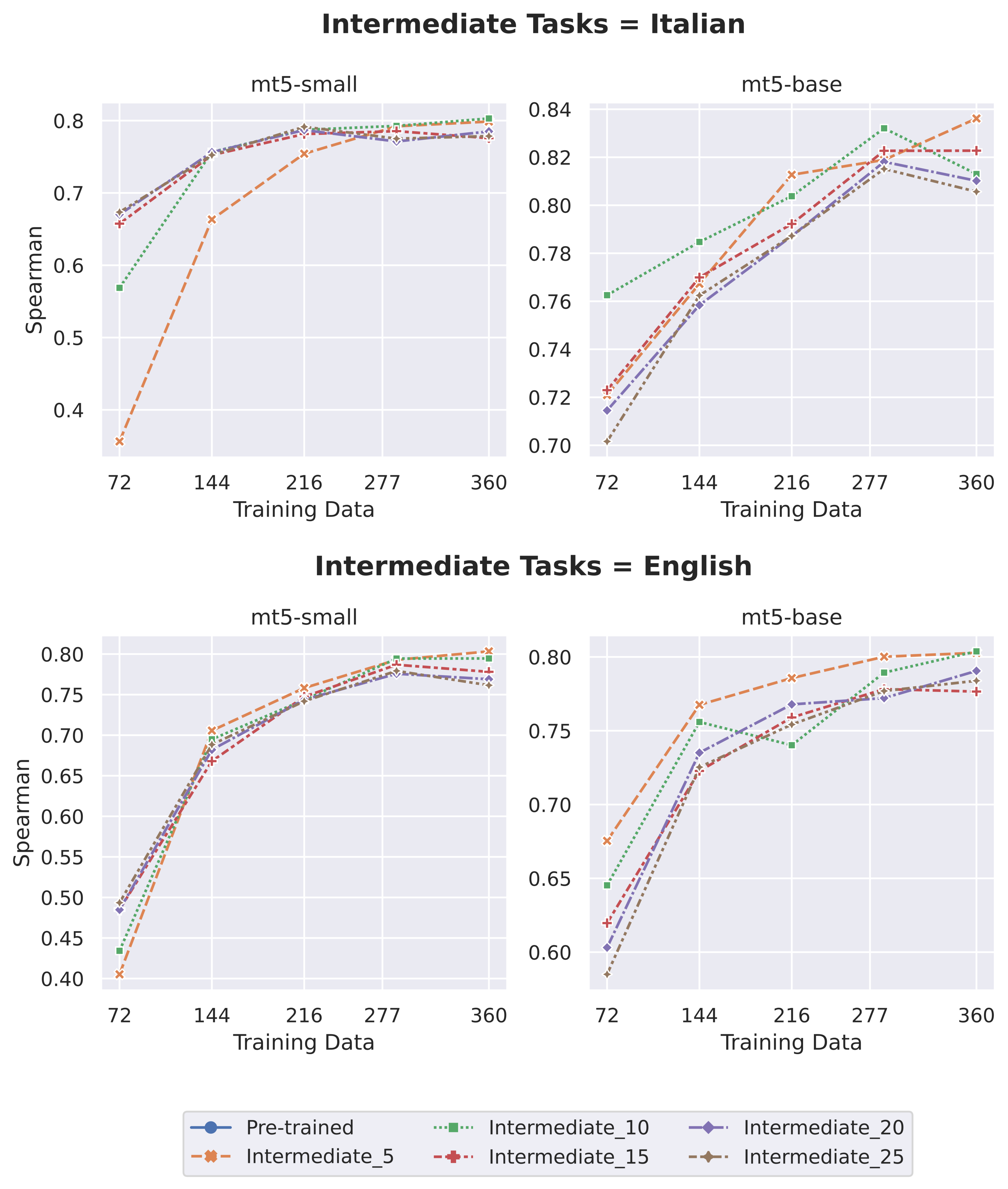}
\caption{Spearman correlation coefficients for the target task on Italian sentences obtained with the multilingual models using pre-trained and LiT5 models.}
\label{fig:m_target_scores_ita}
\end{figure*}

\begin{figure*}[t!]
\centering
\includegraphics[width=0.8\textwidth]{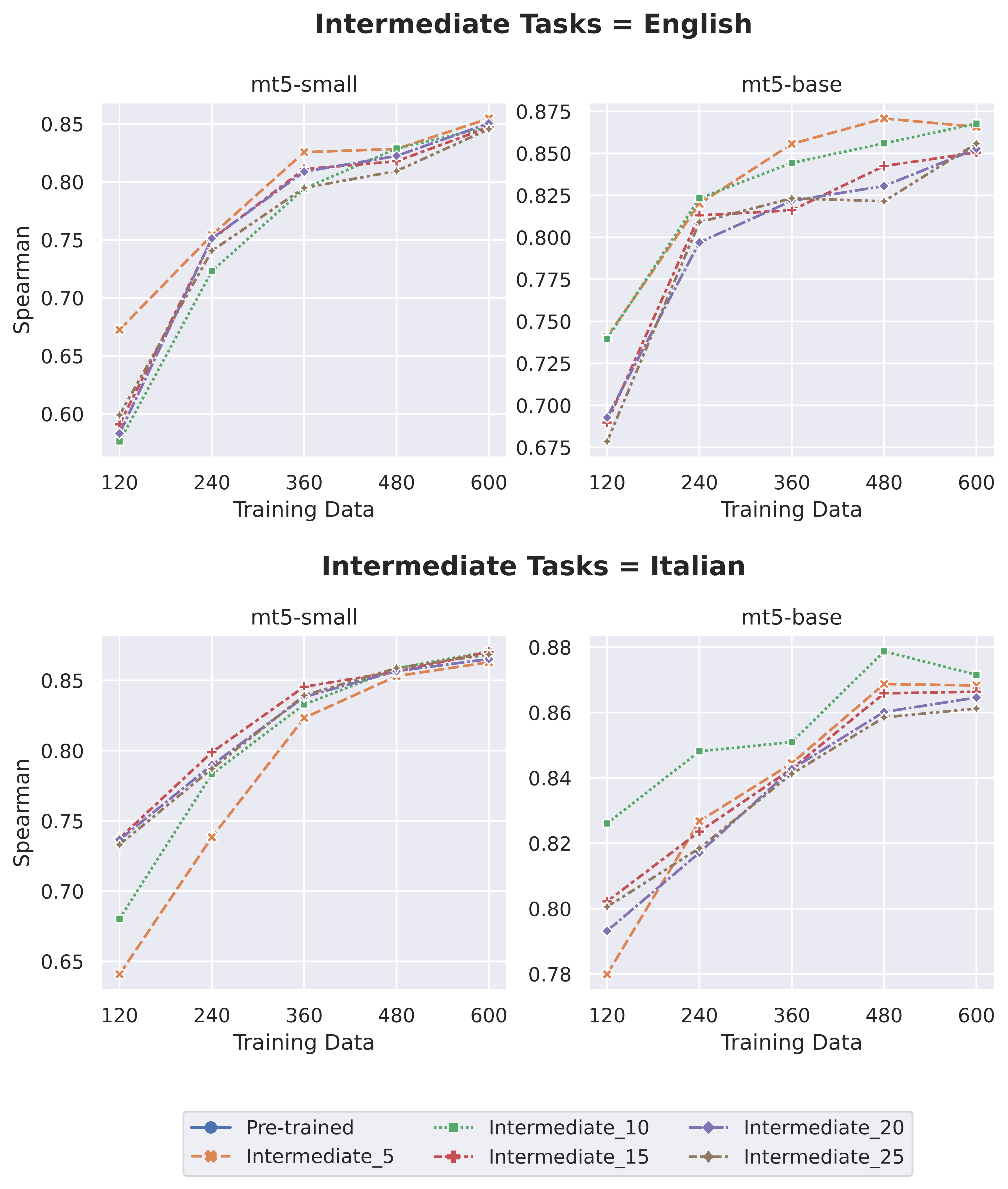}
\caption{Spearman correlation coefficients for the target task on English sentences obtained with the multilingual models using pre-trained and LiT5 models.}
\label{fig:m_target_scores_eng}
\end{figure*}

\begin{table}[t!]
\centering
\small
\begin{tabular}{l|rr|rr}
\hline
\textbf{Model}   & \textbf{LiT5} & \textbf{Diff} & \textbf{LiT5} & 
\textbf{Diff} \\
\hline
& \multicolumn{2}{c|}{\textbf{1/5 Training}} & \multicolumn{2}{c}{\textbf{5/5 Training}} \\
\hline
 \multicolumn{5}{c}{\textbf{Intermediate = Italian, Target = Italian}} \\
\hline
mt5-small             & 25   epochs      & 0.49      & 10  epochs       & 0.37      \\
mt5-base              & 10    epochs     & 0.22      & 5  epochs        & 0.11      \\
\hline
 \multicolumn{5}{c}{\textbf{Intermediate = English, Target = Italian}} \\
 \hline
mt5-small             & 25   epochs      & 0.31      & 5 epochs         & 0.37      \\
mt5-base              & 5    epochs      & 0.13      & 10   epochs      & 0.08     \\
\hline
 \multicolumn{5}{c}{\textbf{Intermediate = English, Target = English}} \\
 \hline
mt5-small             & 5  epochs        & 0.46      & 5    epochs      & 0.15      \\
mt5-base              & 5    epochs      & 0.17      & 10   epochs      & 0.01      \\
\hline
 \multicolumn{5}{c}{\textbf{Intermediate = Italian, Target = English}} \\
 \hline
mt5-small             & 15   epochs      & 0.52      & 10  epochs       & 0.16      \\
mt5-base              & 10    epochs     & 0.25      & 10  epochs       & 0.02      \\
\hline
\end{tabular}
\caption{Performance differences 
between \textit{Pre-trained} and top-performing LiT5 models (trained on Italian/English sentences) in both scenarios of limited (1/5) and full (5/5) data availability.}
\label{table:differences_multi}
\end{table}

We report in Figure \ref{fig:m_target_scores_ita} and \ref{fig:m_target_scores_eng} the results (in terms of Spearman correlation scores) for the target task obtained with the multilingual models on the Italian (Figure \ref{fig:m_target_scores_ita}) and English (Figure \ref{fig:m_target_scores_eng}) datasets. Table \ref{table:differences_multi} reports instead the performance differences between \textit{Pre-trained} and top-performing multilingual LiT5 models (trained on Italian/English sentences) in both scenarios of limited (1/5) and full (5/5) data availability.

\end{document}